\documentclass[conference]{IEEEtran}
\usepackage{cite}
\usepackage{amsmath,amssymb,amsfonts}
\usepackage{algorithmic}
\usepackage{graphicx}
\usepackage{textcomp}
\usepackage{xcolor}

\usepackage{titling}

\usepackage[justification=justified,font=small]{caption}
\usepackage{booktabs}
\usepackage{makecell}

\usepackage{pgfplots}
\usepackage{pgfplotstable}
\pgfplotsset{compat=1.18}
\usepgfplotslibrary{colorbrewer}

\usepackage{pgffor}
\usepackage{calc}

\usepackage{ifthen}

\usepackage{tabularx}
\usepackage{array} 

\newcolumntype{L}[1]{>{\raggedright\arraybackslash}p{#1}}

\usepackage[para,online,flushleft]{threeparttable}

\usepackage{subcaption}
\usepackage{tcolorbox}

\usepackage{stfloats}
\usepackage{dblfnote}
\usepackage{multicol}
\usepackage[hidelinks]{hyperref}

\renewcommand{\footnoterule}{
    \kern 1em
    \hrule width \columnwidth
    \kern 3pt
}

\newcommand\blfootnote[1]{%
  \begingroup
  \renewcommand\thefootnote{}\footnote{#1}%
  \addtocounter{footnote}{-1}%
  \endgroup
}

\def\BibTeX{{\rm B\kern-.05em{\sc i\kern-.025em b}\kern-.08em
    T\kern-.1667em\lower.7ex\hbox{E}\kern-.125emX}}
\begin{document}

\title{Domain-Specific Pretraining of Language Models:\\A Comparative Study in the Medical Field}

\author{Tobias Kerner \\
\textit{Technische Hochschule Ingolstadt}\\
\href{https://orcid.org/0009-0008-5485-1018}{https://orcid.org/0009-0008-5485-1018}
}

\maketitle

\newboolean{includegraphics}
\setboolean{includegraphics}{true}

\begin{abstract}
There are many cases where LLMs are used for specific tasks in a single domain. These usually require less general, but more domain-specific knowledge. Highly capable, general-purpose state-of-the-art language models like GPT-4 or Claude-3-opus can often be used for such tasks, but they are very large and cannot be run locally, even if they were not proprietary. This can be a problem when working with sensitive data.
This paper focuses on domain-specific and mixed-domain pretraining as potentially more efficient methods than general pretraining for specialized language models. We will take a look at work related to domain-specific pretraining, specifically in the medical area, and compare benchmark results of specialized language models to general-purpose language models. 
\end{abstract}


\section{Introduction}

Large Language Models (LLMs) have gained increasing popularity in recent times. It is not unusual to see companies employ them as assistants on websites or employ them directly in workflows like the development of internal IT systems. They also have growing relevance in specialized fields like medicine. LLMs could, for example, assist doctors with diagnoses or explain complicated medical situations to patients in simple language. For models to be used reliably, it is important for them to perform well in the domains of their intended usage. High-performance general-purpose models currently include GPT-4 and claude-3-opus, which can be applied to most tasks with usually good results. 

However, due to their large size and most state-of-the-art (SOTA) general-purpose models being proprietary and only accessible through an API, it is generally not possible to run these LLMs locally. This is a big problem when reliability and privacy are the main concerns. When working with sensitive patient data or with an unreliable internet connection, using an external API is often not feasible. 

A possible solution to this problem would be to use smaller, specialized LLMs that perform well on tasks in a certain domain and whose performance in other domains is irrelevant. Small models benefit from comparatively fast inference times, lower latency, and less expensive training. Their reasonably small size allows them to be hosted on local, consumer-grade hardware. Due to their lower parameter count and limited learning capacity, small models inherently perform worse than large models, given similar training conditions such as the amount and quality of data used for training. Focusing the training on a specific domain and ignoring capabilities in all other domains lets us circumvent this limitation to some degree by not forcing the model to learn irrelevant or out-of-domain information and allowing it to fully focus on memorizing domain-related information. 

But how can we create specialized models, and can they really be as good as large general-purpose models in their domain? To answer these questions, we will take a look at different methods of training LLMs, domain-specific datasets and compare the benchmark results of specialized models to general-purpose models in benchmarks related to their domain. This paper will focus on the medical domain for all examples and comparisons.

\section{Pretraining}

Pretraining is the most fundamental step in creating intelligent and capable LLMs. During this training step, the model learns the structure of natural language and tries to memorize as much of the training data as possible. For causal language models, for example, this happens by inputting token sequences into the model and letting it predict the next token in the sequence. The closer the prediction is to the true next token in the sequence, the smaller the loss. The loss is used to adjust the weights of the model, in order to minimize the loss. When the model learns the contents of the data and improves its predictions, the loss decreases. Pretraining teaches the model general knowledge and language understanding, but not to answer in a chat-like way, as many end-users are familiar with. For this usage, a further pretraining step is often necessary, where the model is trained to reply in a certain format. Usually, finetuning impacts the during pretraining learned knowledge only slightly, since finetuning is a relatively short process compared to pretraining and should not change the models weights a lot.

\subsection{General Pretraining}

General pretraining is considered the default method for training large language models and works well in most cases.
There are many large datasets that can be used for general pretraining without much modification. 
They contain enough varied data to teach the model general language understanding and general knowledge, even in multiple languages.
During training, the model will try to remember as much of the knowledge in the provided data as possible in order to minimize its training loss.
Large models may be able to learn most of the data, but smaller models will struggle with this due to their inherent parameter limit.
This causes them to memorize all information in the data a little bit, but nothing well.
However, in the context of training a specialized model, there are ways to circumvent this issue.

\subsection{Domain-Specific Pretraining}

If the model will be used for tasks in a specific domain only, we can mostly avoid this issue by simply removing everything from the data that is not related to the target domain, leaving us with a domain-specific dataset. 
When training a medical LLM for example, it is to be expected that a lack of training data on poems or music, for example, should not affect its performance on medical-related tasks.
This is only possible if there is enough data left in the domain-specific dataset for the model to get an understanding of natural language from it while not overfitting \cite{pubmedbert}.

The assumption is that the less irrelevant information the model has to learn, the better it can learn the relevant information.
This should make it possible to decrease model size and accelerate training without a drastic decrease in domain-related task performance.
Even though a reduced parameter count will decrease the model's complexity and likely limit its ability to understand very complex data structures, this should not be an issue for simpler tasks or tasks that rely mostly on knowledge. 
An example of this in the medical field would be the task of translating a detailed medical report into simple language for the patient to understand, where the model has to know the concept and meaning of medical terms in order to rephrase the report.

\subsection{Mixed-Domain Pretraining}

Mixed-domain pretraining, also called continued pretraining, can be used if there is not enough data in the domain-specific dataset for the model to gain an understanding of natural language from it.
We can first train the model on a general-purpose dataset. This will enable the model to learn natural language and, as mentioned before, a bit of all the different domains in the data. 
This model with a general understanding of language can be used as a base for continued pretraining on the domain-specific dataset. 
Since this dataset will likely not include many of the things that the model has memorized previously, training on it will result in the model gradually forgetting most of them, which we can ignore when pretraining for domain-specific tasks. 
If physical resources are limited and training an LLM from scratch is not feasible, this method can also be applied using a third-party general LLM as a base for continued pretraining.

continued pretraining on domain-specific data after training on general data can lead to strong improvements of domain-specific models compared to pretraining on general data only. It sets a stronger knowledge foundation for in-domain finetuning. Their experiments show that small models struggle with large amounts of knowledge and perform better on data specialized in a single domain. \cite{gururangan2020dont}

However, if there is enough domain-specific data for domain-specific pretraining without general pretraining before, performance on domain-specific tasks may be better with domain-specific pretraining than mixed-domain pretraining. While transfer-learning by pretraining on general data is useful for models specialized on domains with low amounts of data, it could be harmful for domain-specific models like medical models. This may be because biomedical texts are substantially different from general texts or different domains, so the transfer-learning might actually have a negative effect. \cite{pubmedbert}

This negative effect may arise from the embedding layer and the first few layers in the model not benefiting from an early focus on domain-related data, which could enable more efficient processing of complex domain-related data in further layers.

\section{Datasets}
There are many publicly available datasets that can be used for pretraining LLMs, including general datasets as well as domain-specific ones.
They are created by collecting data from web scraping, books, articles, or other text-based sources and then processing / cleaning that data. 
Their sizes can range from a few kilobytes to terabytes.

\subsection{General Datasets}

General-purpose datasets contain various types of content and tend to be a lot larger than domain-specific ones. Table \ref{table_general_datasets} shows some of the largest open-source datasets currently available for pretraining. Their extensive size allows the model to gain a solid understanding of language, as well as broad general knowledge.

\ifthenelse{\boolean{includegraphics}}{
    \begin{table}[!th]
        \centering
        \resizebox{\columnwidth}{!}{
        \begin{threeparttable}
        \caption{Pretraining Datasets}
        \label{table_general_datasets}
        \small
        \centering
        \begin{tabularx}{0.75\textwidth}{@{} l|c|c|c|c|L{\columnwidth} @{}}
        \toprule
        \multicolumn{1}{c|}{Dataset} & Type & Size (GiB) & Language & Year & Description  \\ 
        \midrule
        CommonCrawl\cite{commoncrawl} & Web & 391,168 & misc & 2024 & 2.7B webpages, June 2024 crawl\newline (first release in 2008) \\
        The Stack\cite{gao2020pile800gbdatasetdiverse} & Code & 3,136 & misc & 2022 & Source codes in 30 different\newline programming languages \\
        The Pile\cite{kocetkov2022stack3tbpermissively} & Mixed & 825 & en & 2020 & Consisting of 22 sub-sets related\newline to various domains and tasks \\
        RefinedWeb\cite{NEURIPS2023_fa3ed726} & Web & 2,800 & en & 2023 & Based on cleaned and refined data\newline from Common Crawl \\
        \bottomrule
        \end{tabularx}
        \end{threeparttable}
        }
    \end{table}
}{}

It is to note that CommonCrawl-data is used partially in many other datasets, such as RefinedWeb \cite{NEURIPS2023_fa3ed726}. It contains an extensive collection of raw scraped websites, which can be filtered, cleaned, and processed for the creation of new, refined datasets. Figure \ref{fig:commoncrawl_processing} shows how the creators of RefinedWeb processed CommonCrawl to create their dataset. A large part of the raw data gets filtered out during language identification, during which they keep only English texts, and removal of duplicates. \cite{NEURIPS2023_fa3ed726}

\ifthenelse{\boolean{includegraphics}}{
    \begin{figure}[t]
        \centering
        \includegraphics[width=\columnwidth]{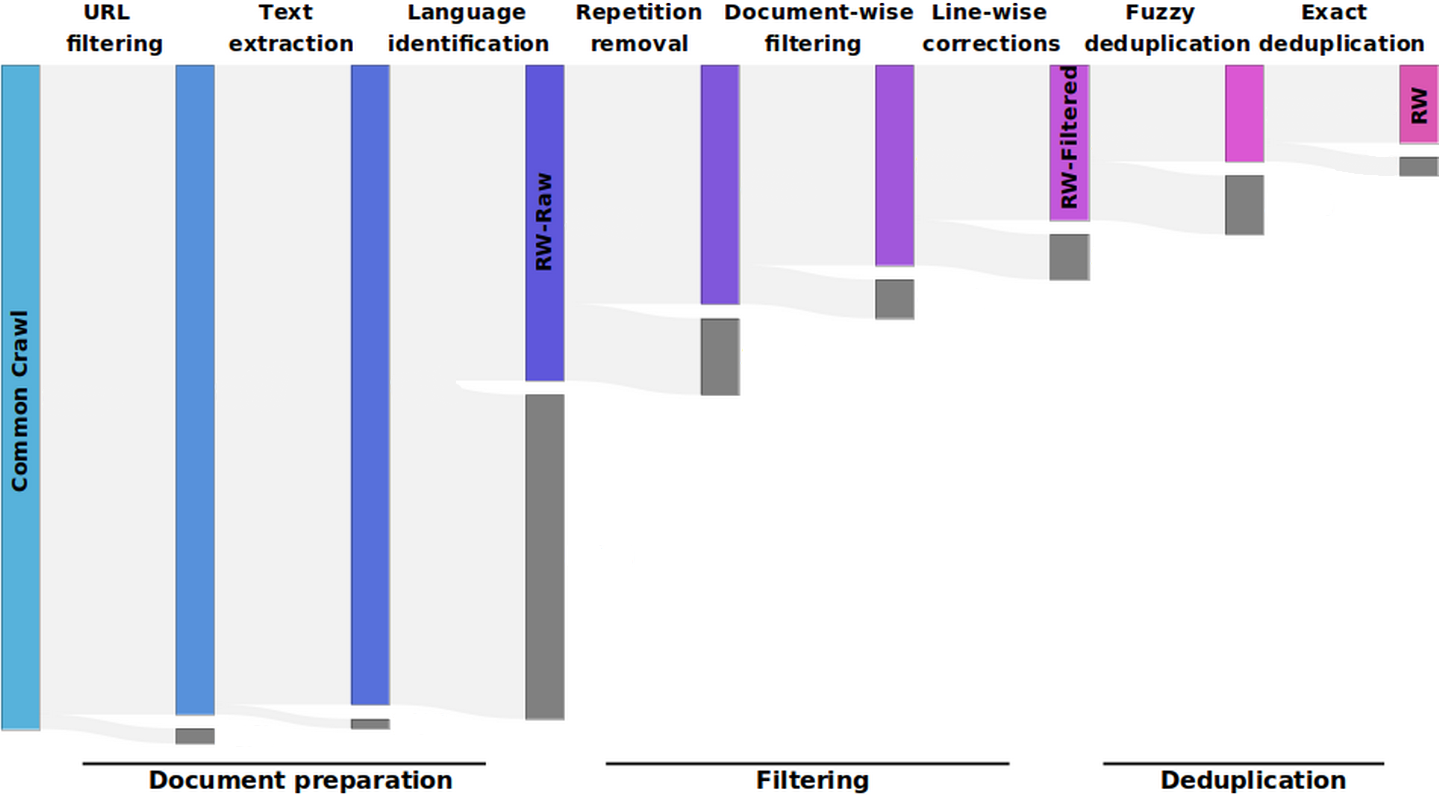}
        \caption{Processing of CommonCrawl Dataset for RefinedWeb, taken from \cite{NEURIPS2023_fa3ed726} and edited for simplification.}
        \label{fig:commoncrawl_processing}
    \end{figure}
}{}

\subsection{Medical Datasets Overview}

There are publicly available domain-specific datasets as well. They can be used to further refine a model's knowledge and understanding in a specific area. Table \ref{table_medical_datasets} shows select open-source medicine-related datasets, some of which will also be used for benchmarking in the next section of this paper. These are a lot smaller than the general datasets in table \ref{table_general_datasets} due to the domain limitation and the requirement for high-quality content. The quality requirement for domain-specific datasets is higher since they tend to be used to refine a base-model's capabilities in a specific domain.

Some of the well-known medical datasets include the following:
MeDAL consists of PubMed-abstracts with around 3 abbreviations per abstract. The authors goal was to create a dataset that teaches LLMs to handle medical-related abbreviations better \cite{wen-etal-2020-medal}.
MedDialog contains English and Chinese dialogues. The English dialogues have exactly 2 utterances per dialogue, with an average length of 86.5 tokens per utterance. The Chinese dialogues are longer, with an average of 3.3 utterances per dialogue up to a maximum of 198 utterances in one dialogue and an average of 55.6 tokens per utterance \cite{zeng-etal-2020-meddialog}.
MedQA consists of English and Chinese question-answer pairs, collected from medical board examinations in the USA, China, and Taiwan \cite{app11146421}.
PubMedQA contains question-answer pairs that are based on PubMed abstracts. It is split into 1k expert-annotated, 61.2k unlabeled, and 211.3k artificially generated QA-pairs \cite{jin-etal-2019-pubmedqa}.
MedMCQA is designed to imitate real-world medical entrance exam questions. It contains questions in natural language with four possible answers, one correct answer, and an explanation for each \cite{pal2022medmcqa}.

\ifthenelse{\boolean{includegraphics}}{
    \begin{table}[ht]
        \centering
        \resizebox{\columnwidth}{!}{
        \begin{threeparttable}
        \caption{Overview and comparison of medical datasets}
        \label{table_medical_datasets}
        \small
        \centering
        \setlength{\tabcolsep}{4.5pt}
        \begin{tabular}{@{} l *4{|c} @{}}
        \toprule
        \multicolumn{1}{c|}{Dataset} & Type & Entries & Language & Release Year  \\ 
        \midrule
        MeDAL & Articles & 14.4m & en & 2020 \\
        MedDialog & Dialogues & 3.4m, 257.3k & en, ch & 2020 \\
        MedQA & QA & 12.7k, 48.4k & en, ch & 2021  \\
        PubMedQA & QA & 1.0k, 61.2k, 211.3k & en & 2019 \\
        MedMCQA & multiple-choice-QA & 193.1k & en & 2022 \\
        \bottomrule
        \end{tabular}
        \end{threeparttable}
        }
    \end{table}
}{}

\subsection{Creating Domain-Specific Datasets}

One of the biggest problems regarding training language models is the amount of high-quality data available. For large-scale domain-specific training, we will likely need more domain-related data than is available. For some domains, there may not even be a publicly available dataset yet, or it does not fulfill some requirements. In this case, we would have to create our own dataset. There are several steps we have to take for creating our own domain-specific dataset, including the acquisition of raw data, filtering and cleaning it, and processing it into a format suitable for training.

For acquiring raw data, we can either collect it ourselves by means of webscraping, using a dump of already scraped data such as CommonCrawl, or collecting multiple datasets that are at least slightly related to our domain. Large-scale webscraping is a time- and resource-intensive task, so it is recommended to consider using data from CommonCrawl instead. The team behind it releases regular updates to the dataset with newly scraped websites. Due to its extensive size, it is more than likely to contain data related to our domain. \cite{commoncrawl}

Assuming we chose to use CommonCrawl as the source for our raw data, we can now filter it for data related to our domain, which would be medicine in our case. One way to do this is by iterating through the scraped websites and looking for medicine-related keywords, saving the websites containing them separately. Other methods to filter for relevant data include Named Entity Recognition, topic modeling (using techniques such as Latent Dirichlet Analysis) or using a pretrained language model to tell us if a text is of medical nature or not. These methods can yield better results than simple keyword-matching, but are more complex. The best filtering can likely be achieved by using a combination multiple methods.

After collecting enough in-domain data, it will have to be cleaned. This means removing duplicate sites, as well as noise and irrelevant data embedded inside the relevant text. Noise can, for example, be caused by websites inserting advertisements in the middle of their articles. Keeping this would be harmful to our training since it would likely distract or confuse the model and hence has to be removed.

This cleaned data containing medical text could be used for training now, but we can further refine it by creating question-answer pairs for it. This can be especially useful for teaching causality and the relationship between concepts and entities in the text. For this, there are tools like Augmentoolkit. Provided with data in text format, it splits the data into chunks and employs an LLM to extract suitable questions and the corresponding answers from these chunks \cite{augmentoolkit}. Due to the nature of language models, it is not guaranteed that all extracted question-answer pairs are correct, even assuming that the provided data was flawless. But this disadvantage would likely be negligible in comparison to the potential benefit of training on data in this format. Of course, we can first train our LLM on the normal cleaned data and then use this refined question-answer data later for finetuning.

\section{Performance of specialized LLMs}

As mentioned previously, specialized LLMs are usually smaller than general-purpose LLMs. They are trained on less out-of-domain data, speeding up training time and reducing compute cost. In the following, we will take a look at medical LLMs trained with domain-specific pretraining and mixed-domain pretraining and compare their performance to general-purpose LLMs of equal or larger size in medical-related tasks.

\subsection{Domain-Specific Pretrained Models}

A well-known medicine-focused model is PubMedBERT. It is based on the BERT architecture \cite{bert} and trained on 14 million PubMed-abstracts with a total length of 3.1 billion words. The model was trained from scratch, only using the BERT architecture but not the pretrained weights of the original BERT model. Its tokenizer uses vocabulary newly generated with the WordPiece algorithm, based on the PubMed-data that was also used for training. \cite{pubmedbert}

BioMedLM is a 2.7B model trained exclusively on biomedical text, including PubMed-abstracts and full articles.
When finetuned to reply in the format required by a specific task, BioMedLM is competitive with much larger models, as shown by its score of 57.3\% on MedMCQA (dev) and 69.0\% on the MMLU Medical Genetics exam. Even though BioMedLM only has 2.7B parameters, it beats GPT-3.5 with 175B parameters on MedMCQA and is on par with it in the MedQA and PubMedQA benchmarks. However, it seems to struggle against larger models in MMLU general medical knowledge benchmarks. \cite{bolton2024biomedlm}

Apollo is a family of small LLMs ranging from 0.5B to 7B parameters in size. It is trained using a custom multilingual medical dataset called ApolloCorpora. This dataset is based on books, papers, and encyclopedias related to the medical domain, as well as doctor-patient dialogues with a total of 2.5B training tokens. \cite{wang2024apollo}

\subsection{Mixed-Domain Pretrained}

HEAL is based on LLaMA2-13B, applying continued pretraining to the weights of the original model with 14.89B tokens of medical data.
It surpasses GPT-4 on PubMedQA, with a score closely behind Med-PaLM-2. It seems to struggle on the MedQA benchmark, with only a small improvement over the base LLaMA2-13B and quite behind another LLaMA-13B-based model. It is worth noting that its score of 47.2 on MedQA is around the same level as GPT-3.5. \cite{yuan-etal-2024-continued, nori2023capabilities}

\subsection{Benchmark Comparison}
    \begin{figure}[t]
        \centering
        \begin{threeparttable}
        \caption{Comparison of models on medical benchmarks, sorted by model size. The number in brackets behind the score represents the evaluation method (x-shot). f stands for finetune, the model was finetuned for this task. if the reference paper does not provide the evaluation method or specifies 'few-shot', there brackets will contain a '?'.}
        \label{table_comparison_models}
        \small
        \centering
        \setlength{\tabcolsep}{3pt}
        \begin{tabular}{@{} l *7{c} @{}}
        \toprule
        \multicolumn{1}{c}{Model} 
                  & Size & MedMCQA  & PubMedQA & MedQA    & MMLU      \\ 
        \midrule
        Apollo    & 0.5B & 37.8 (3) & -        & -        & 45.9 (3)  \\
        Apollo    & 1.8B & 45.0 (3) & -        & -        & 49.1 (3)  \\
        BioMedLM  & 2.7B & 57.3 (f) & 74.4 (f) & 50.3 (f) & 63.1 (f)  \\
        LLaMA 2   & 7B   & 36.6 (3) & -        & -        & 40.1 (3)  \\
        Apollo    & 7B   & 58.2 (3) & -        & -        & 71.9 (3)  \\
        LLaMA 2   & 13B  & 37.4 (f) & 76.4 (3) & 45.5 (3) & -         \\
        HEAL      & 13B  & -        & 78.4 (3) & 47.2 (3) & -         \\
        LLaMA 2   & 70B  & 48.3 (3) & -        & -        & 64.6 (3)  \\
        GPT-3.5   & 175B & 51.0 (0) & 71.6 (0) & 44.6 (0) & 70.6 (0)  \\
        GPT-4     & 1.7T\tnote{1} & 73.4 (0) & 80.4 (0) & 81.4 (0) & 92.6 (0)  \\

        \bottomrule
        \end{tabular}
        \begin{tablenotes}
            \footnotesize
            \item[1] rumoured parameter count \cite{substackGPT4Details}
        \end{tablenotes}
        \end{threeparttable}

\end{figure}

As shown in table \ref{table_comparison_models}, the specialized LLMs perform well in in-domain tasks with respect to their size when compared to not domain-specifically trained models. Especially BioMedLM shows strong performance, beating GPT-3.5 in MedMCQA, PubMedQA, and MedQA with only 2.7B parameters.
The Apollo models perform good as well, with Apollo-7B improving upon BioMedLM's MMLU score, slightly beating GPT-3.5. This is surprising, considering it was trained on 2.5B tokens of data only, compared to BioMedLM's 300B tokens. With a size of just 0.5B parameters, Apollo-0.5B reaches solid scores in MedMCQA and MMLU, beating the about 14 times larger LLaMA-2-7B. Interestingly, both GPT-3.5 and GPT-4 perform worse on PubMedQA with 5-shot than with 0-shot, having a score of 60.2 and 77.4 with 5-shot, respectively \cite{nori2023capabilities}. 
\blfootnote{Data for table \ref{table_comparison_models}: The parts of MMLU used for calculating the score for all Apollo models, LLaMA-2-70B, and LLaMA-2-7B were specified as 'medical-related parts of MMLU' \cite{wang2024apollo}.
For BioMedLM, GPT-3.5, and GPT-4, the MMLU score is the average of the parts 'Clinical Knowledge', 'Professional Medicine', 'College Biology', and 'Medical Genetics'. The scores of GPT-4 are for the base-version of GPT-4, not to be confused with the publicly available version of GPT-4 that was finetuned for more safe and responsible answers and generally performs worse on medical benchmarks. The MedQA score for GPT-3.5 and GPT-4 is based on 'US (5-option)'. \cite{nori2023capabilities}
The LLaMA-2-13B MedMCQA score was taken from \cite{10.1093/jamia/ocae045}.
}

\pgfplotstableread{
    tokenstrained avgscorepubmedqa paramnum color label xshift yshift
    13000 80.4 1760 gray GPT-4 0 -5
    300 71.6 175 gray GPT-3.5 0 -3
    2000 76.4 13 gray Llama-2-13B 0 -6
    300 74.4 2.7 green BioMedLM 0 -1
    15 78.4 13 orange HEAL 0 -2
}\datatablepubmedqa

\pgfplotstablegetrowsof{\datatablepubmedqa}
\newcommand{\nrowspubmedqa}{\pgfplotsretval}
\pgfmathsetmacro{\rowstoiteratepubmedqa}{\nrowspubmedqa-1}

\pgfplotstableread{
    tokenstrained avgscoremedmcqa paramnum color label xshift yshift
    13000 73.4 1760 gray GPT-4 0 -5
    300 51.0 175 gray GPT-3.5 0 -2
    2000 48.3 70 gray Llama-2-70B -2 2
    2000 37.4 13 gray Llama-2-13B 0 -5
    2000 36.6 7 gray Llama-2-7B -5 5
    300 57.3 2.7 green BioMedLM -2 5
    2.5 58.2 7 green Apollo-7B 0 -2
    2.5 45.0 2.7 green Apollo-2.7B 0 1
    2.5 37.8 0.5 green Apollo-0.5B 0 1
}\datatablemedmcqa

\pgfplotstablegetrowsof{\datatablemedmcqa}
\newcommand{\nrowsmedmcqa}{\pgfplotsretval}
\pgfmathsetmacro{\rowstoiteratemedmcqa}{\nrowsmedmcqa-1}

\ifthenelse{\boolean{includegraphics}}{
    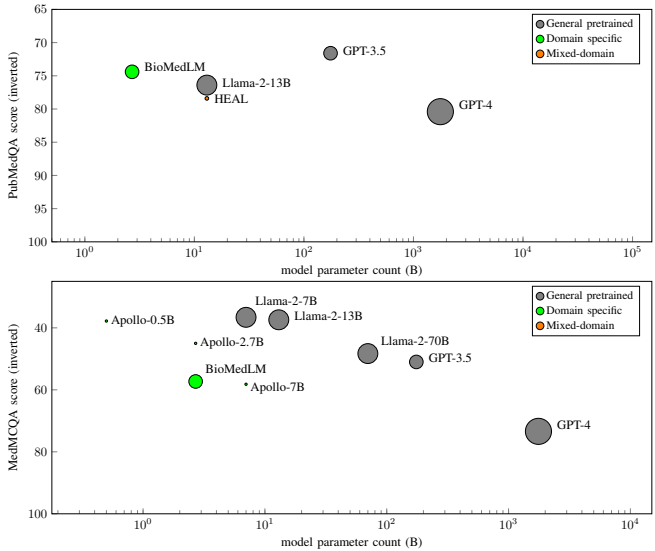
\begin{figure}[t]
        \resizebox{\columnwidth}{!}{
        \begin{subfigure}[b]{\textwidth}
        \begin{tikzpicture}
            \begin{axis}[
                width=\textwidth,
                height=8cm,
                xlabel={model parameter count (B)},
                ylabel={PubMedQA score (inverted)},
                xmin=0.5, xmax=150000,
                ymin=65, ymax=100,
                xmode=log,
                ymode=normal,
                log basis y={10},
                y dir=reverse,
                xtick pos=bottom,
                ytick pos=left,
                legend style={
                    font=\small,
                    anchor=north east,
                    fill=none,
                    align=left,
                    nodes={anchor=west},
                },
            ]
            \addlegendimage{only marks, mark=*, mark size=3pt, color=gray, mark options={draw=black, line width=0.1pt}}
            \addlegendentry{General pretrained}
            \addlegendimage{only marks, mark=*, mark size=3pt, color=green, mark options={draw=black, line width=0.1pt}}
            \addlegendentry{Domain specific}
            \addlegendimage{only marks, mark=*, mark size=3pt, color=orange, mark options={draw=black, line width=0.1pt}}
            \addlegendentry{Mixed-domain}
            \pgfplotsinvokeforeach {0, 1, ..., \rowstoiteratepubmedqa} {
                \pgfplotstablegetelem{#1}{[index]2}\of\datatablepubmedqa
                \edef\xcoord{\pgfplotsretval}
                \pgfplotstablegetelem{#1}{[index]1}\of\datatablepubmedqa
                \pgfmathsetmacro{\ycoord}{\pgfplotsretval}
                \pgfplotstablegetelem{#1}{[index]0}\of\datatablepubmedqa
                \pgfmathsetmacro{\circlesize}{(log10((\pgfplotsretval)/5)*3}
                \pgfplotstablegetelem{#1}{[index]3}\of\datatablepubmedqa
                \edef\circlecolor{\pgfplotsretval}
                \pgfplotstablegetelem{#1}{[index]4}\of\datatablepubmedqa
                \edef\labeltextin{\pgfplotsretval}
                \pgfplotstablegetelem{#1}{[index]5}\of\datatablepubmedqa
                \pgfmathsetmacro{\labelshiftright}{\pgfplotsretval+\circlesize+1}
                \pgfplotstablegetelem{#1}{[index]6}\of\datatablepubmedqa
                \pgfmathsetmacro{\labelshiftup}{\pgfplotsretval+\circlesize}
                \edef\plotcmd{\noexpand\addplot[only marks, mark=*, mark size=\circlesize, color=\circlecolor, mark options={draw=black, line width=0.1pt}] coordinates {(\xcoord, \ycoord)} node [right, black, yshift=\labelshiftup, xshift=\labelshiftright] {\labeltextin};}
                \plotcmd
            }
            \end{axis}
        \end{tikzpicture}
        \label{benchmark_comparison_pubmedqa}
        \end{subfigure}
        }

        \resizebox{\columnwidth}{!}{
        \begin{subfigure}[b]{\textwidth}
        \begin{tikzpicture}
            \begin{axis}[
                width=\textwidth,
                height=8cm,
                xlabel={model parameter count (B)},
                ylabel={MedMCQA score (inverted)},
                xmax=15000,
                ymin=25, ymax=100,
                xmode=log,
                ymode=normal,
                log basis y={10},
                y dir=reverse,
                xtick pos=bottom,
                ytick pos=left, 
                legend style={
                    font=\small,
                    anchor=north east,
                    fill=none,
                    align=left,
                    nodes={anchor=west},
                },
            ]
            \addlegendimage{only marks, mark=*, mark size=3pt, color=gray, mark options={draw=black, line width=0.1pt}}
            \addlegendentry{General pretrained}
            \addlegendimage{only marks, mark=*, mark size=3pt, color=green, mark options={draw=black, line width=0.1pt}}
            \addlegendentry{Domain specific}
            \addlegendimage{only marks, mark=*, mark size=3pt, color=orange, mark options={draw=black, line width=0.1pt}}
            \addlegendentry{Mixed-domain}
            \pgfplotsinvokeforeach {0, 1, ..., \rowstoiteratemedmcqa} {
                \pgfplotstablegetelem{#1}{[index]2}\of\datatablemedmcqa
                \edef\xcoord{\pgfplotsretval}
                \pgfplotstablegetelem{#1}{[index]1}\of\datatablemedmcqa
                \pgfmathsetmacro{\ycoord}{\pgfplotsretval}
                \pgfplotstablegetelem{#1}{[index]0}\of\datatablemedmcqa
                \pgfmathsetmacro{\circlesize}{(log10((\pgfplotsretval)/5)*3}
                \pgfplotstablegetelem{#1}{[index]3}\of\datatablemedmcqa
                \edef\circlecolor{\pgfplotsretval}
                \pgfplotstablegetelem{#1}{[index]4}\of\datatablemedmcqa
                \edef\labeltextin{\pgfplotsretval}
                \pgfplotstablegetelem{#1}{[index]5}\of\datatablemedmcqa
                \pgfmathsetmacro{\labelshiftright}{\pgfplotsretval+\circlesize+1}
                \pgfplotstablegetelem{#1}{[index]6}\of\datatablemedmcqa
                \pgfmathsetmacro{\labelshiftup}{\pgfplotsretval+\circlesize}
                \edef\plotcmd{\noexpand\addplot[only marks, mark=*, mark size=\circlesize, color=\circlecolor, mark options={draw=black, line width=0.1pt}] coordinates {(\xcoord, \ycoord)} node [right, black, yshift=\labelshiftup, xshift=\labelshiftright] {\labeltextin};}
                \plotcmd
            }
            \end{axis}
        \end{tikzpicture}
        \label{benchmark_comparison_medmcqa}
        \end{subfigure}
        }
        \caption{Comparison of model scores on PubMedQA and MedMCQA. Circle size is logarithmically proportional to the number of tokens the model was trained with. Number of training tokens for general models taken from \cite{brown2020languagemodelsfewshotlearners, substackGPT4Details, touvron2023llama2openfoundation}.}
        \label{fig:benchmark_comparison}
    \end{figure}
}{}

\subsection{Comparing Model-Size to Benchmark Score}

\ifthenelse{\boolean{includegraphics}}{
    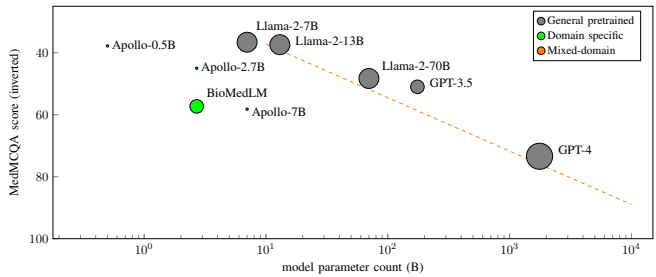
\begin{figure}[t]
    \resizebox{\columnwidth}{!}{
    \begin{subfigure}[b]{\textwidth}
    \begin{tikzpicture}
        \begin{axis}[
            width=\textwidth,
            height=8cm,
            xlabel={model parameter count (B)},
            ylabel={MedMCQA score (inverted)},
            xmax=15000,
            ymin=25, ymax=100,
            xmode=log,
            ymode=normal,
            log basis y={10},
            y dir=reverse,
            xtick pos=bottom, 
            ytick pos=left,
            legend style={
                font=\small,
                anchor=north east,
                fill=none,
                align=left,
                nodes={anchor=west},
            },
        ]
        \addlegendimage{only marks, mark=*, mark size=3pt, color=gray, mark options={draw=black, line width=0.1pt}}
        \addlegendentry{General pretrained}
        \addlegendimage{only marks, mark=*, mark size=3pt, color=green, mark options={draw=black, line width=0.1pt}}
        \addlegendentry{Domain specific}
        \addlegendimage{only marks, mark=*, mark size=3pt, color=orange, mark options={draw=black, line width=0.1pt}}
        \addlegendentry{Mixed-domain}
        \pgfplotsinvokeforeach {0, 1, ..., \rowstoiteratemedmcqa} {
            \pgfplotstablegetelem{#1}{[index]2}\of\datatablemedmcqa
            \edef\xcoord{\pgfplotsretval}
            \pgfplotstablegetelem{#1}{[index]1}\of\datatablemedmcqa
            \pgfmathsetmacro{\ycoord}{\pgfplotsretval}
            \pgfplotstablegetelem{#1}{[index]0}\of\datatablemedmcqa
            \pgfmathsetmacro{\circlesize}{(log10((\pgfplotsretval)/5)*3}
            \pgfplotstablegetelem{#1}{[index]3}\of\datatablemedmcqa
            \edef\circlecolor{\pgfplotsretval}
            \pgfplotstablegetelem{#1}{[index]4}\of\datatablemedmcqa
            \edef\labeltextin{\pgfplotsretval}
            \pgfplotstablegetelem{#1}{[index]5}\of\datatablemedmcqa
            \pgfmathsetmacro{\labelshiftright}{\pgfplotsretval+\circlesize+1}
            \pgfplotstablegetelem{#1}{[index]6}\of\datatablemedmcqa
            \pgfmathsetmacro{\labelshiftup}{\pgfplotsretval+\circlesize}
            \edef\plotcmd{\noexpand\addplot[only marks, mark=*, mark size=\circlesize, color=\circlecolor, mark options={draw=black, line width=0.1pt}] coordinates {(\xcoord, \ycoord)} node [right, black, yshift=\labelshiftup, xshift=\labelshiftright] {\labeltextin};}
            \plotcmd
        }
        \addplot [
            color=orange,
            domain=10:10000,
            samples=2,
            dashed
        ] {9.226*ln(24.64*x+665.24)-25.59}; 
        \end{axis}
    \end{tikzpicture}
    \end{subfigure}
    }
    \caption{Comparison of model scores on MedMCQA with a trend line based on general pretrained model data. Circle size is logarithmically proportional to the number of tokens the model was trained with. Trend line function: 9.226*ln(24.64*x+665.24)-25.59} 
    \label{fig:benchmark_comparison_line}
    \end{figure}
}{}

Given a high-quality dataset, a model's performance seems to be directly related to its parameter count. According to our visualization in table \ref{fig:benchmark_comparison}, especially the MedMCQA benchmark, we can see a strong relationship between model size and benchmark score for the general pretrained LLMs. 
In \ref{fig:benchmark_comparison_line}, we can see the MedMCQA benchmark plot with a line based on the general pretrained model scores. The domain-specific and mixed-domain pretrained models seem to be able to 'escape' this line and achieve a better score compared to their size. Based on this figure, BioMedLM has the best performance-to-parameter-count relationship. This is likely possible due to the large quantity of medical data it was trained on, enabling it to saturate its 2.7B weights with medical knowledge. It is possible that this relationship is not as strong in different benchmarks, but this example is an indication that it might be at least similar.

\subsection{Further Resource-Optimization}

In the context of our initial question, we asked if LLMs small enough to be run locally on consumer-grade hardware can match the performance of large, general-purpose models that cannot be run locally in certain tasks. We should be able to run a model with 7B parameters on local hardware with around 16GB of VRAM, assuming the weights are saved as (b)float16. But a further reduction in memory requirements and increase in inference speed are possible by quantization of the model's weights. 

Current popular quantization formats include gguf, exl2, awq, and more. Gguf allows for some of the model's layers to be loaded into cpu-memory if the model does not fully fit into gpu-memory, at the cost of inference speed due to slower inference on the cpu. 
While the original Apollo-7B with (b)float16 weights would require slightly less than 16GB of gpu-memory due to each of the slightly more than 7 billion weights each taking up 16 bits / 2 bytes, the Q8\_K quantized version only takes up 9.2GB. Most commonly used quantizations go down to Q4, which stores each weight with an average of 4 bits. The fewer bits are used to store weights, the stronger the degradation of the model's performance becomes. Below 4-bit quantization, especially for smaller models (7B and below), the precision is often too low to be of reliable use. The Q4\_K\_S and Q4\_K\_M quantized versions of Apollo-7B take up only 5.1 and 5.4 GB of memory, respectively. The difference between the K\_S and the K\_M versions is that the K\_M version employs a less strict quantization, with some more performance-impacting layers being kept at higher quanitization levels, resulting in a slightly higher memory requirement. Some weights are not quantized at all in the gguf model, such as those of normalization layers. Because of this, we do not see a perfect 50\% reduction in memory requirements when switching from the 16-bit model to the 8-bit model. Although the gguf-quantized models huggingface-page does not specifically state the precision changes of the model in the quantized versions, the 8-bit gguf quantization is generally considered to have a precision very similar to the original model, making it a viable and, in some cases, even better-suited option than the 16-bit model. \cite{hfapollogguf}

\section{Summary}

Depending on the use case and available resources, domain-specific and mixed-domain pretraining can be a viable and preferable alternative to general pretraining.
With a low amount of in-domain data available, mixed-domain pretraining on a large general dataset and then switching to the small in-domain dataset will likely result in a better-performing model than training on a small domain-specific dataset only.
These specialized models can have a much lower parameter count but still perform impressively well on in-domain tasks compared to larger, general pretrained models.
This makes it possible to run them on consumer-grade hardware, enabling independent local inference.
Therefore, domain-specific pretrained LLMs can be a good alternative to api-based SOTA general-purpose models in some cases, such as supporting symptom analysis and rephrasing documents in a medical setting.

\bibliographystyle{IEEEtran}
\bibliography{ref}

\end{document}